\documentclass[10pt,twocolumn,letterpaper]{article}

\usepackage[pagenumbers]{cvpr} 
\usepackage{verbatim}
\usepackage[dvipsnames]{xcolor}

\usepackage{makecell}
\usepackage{multirow}

\definecolor{cvprblue}{rgb}{0.21,0.49,0.74}
\usepackage[pagebackref,breaklinks,colorlinks,citecolor=cvprblue]{hyperref}

\def\paperID{12415} 
\def\confName{CVPR}
\def\confYear{2024}

\title{BIM: Block-Wise Self-Supervised Learning with Masked Image Modeling}

\author{Yixuan Luo\\
University of Rochester\\
\and
Mengye Ren\\
New York University\\
\and
Sai Qian Zhang\\
New York University\\
}

\begin{document}
\maketitle
\begin{abstract}

Like masked language modeling (MLM) in natural language processing, masked image modeling (MIM) aims to extract valuable insights from image patches to enhance the feature extraction capabilities of the underlying deep neural network (DNN).
Contrasted with other training paradigms like supervised learning and unsupervised contrastive learning, masked image modeling (MIM) pretraining typically demands significant computational resources in order to manage large training data batches (e.g., 4096). The significant memory and computation requirements pose a considerable challenge to its broad adoption. To mitigate this, we introduce a novel learning framework, termed~\textit{Block-Wise Masked Image Modeling} (BIM). This framework involves decomposing the MIM tasks into several sub-tasks with independent computation patterns, resulting in block-wise back-propagation operations instead of the traditional end-to-end approach. Our proposed BIM maintains superior performance compared to conventional MIM while greatly reducing peak memory consumption. Moreover, BIM naturally enables the concurrent training of numerous DNN backbones of varying depths. This leads to the creation of multiple trained DNN backbones, each tailored to different hardware platforms with distinct computing capabilities. This approach significantly reduces computational costs in comparison with training each DNN backbone individually. Our framework offers a promising solution for resource constrained training of MIM.
\end{abstract}    
\section{Introduction}
\label{sec:intro}






To enhance DNN performance, many current methodologies still rely heavily on supervised training paradigms, which demand a substantial amount of manual annotations.
Recent advancements in self-supervised learning (SSL) offer an alternative solution with superior training accuracy and data efficiency. SSL leverages the inherent structures and patterns present in images, enabling the application of learned representations to specific target tasks using techniques like fine-tuning or linear classification. Among the diverse range of SSL strategies posited to date, Masked Image Modeling (MIM)-based approach~\cite{He2022MAE,xie2021simmim} has stood out due to its superior accuracy performance over the conventional supervised frameworks. 
MIM supervises the network by reconstructing occluded image patches by utilizing the visual representations from their visible counterparts. This paradigm intrinsically drives the encoders of Vision Transformer (ViT) to capture important features and patterns while discarding irrelevant or noisy information within an image.

However, MIM approaches have their downsides, 
as they necessitate substantial computational resources for processing large training data batches (e.g., 4096) with long training iterations (e.g., 800 epochs)~\cite{He2022MAE}. This results in significantly higher memory usage and computational cost compared to conventional supervised learning, making the research and development in MIM prohibitively expensive in many scenarios.



One way to attain efficient MIM training with reduced memory usage and computational requirements is to decompose the large DNN into smaller ones and train them separately in parallel.
It has been shown in the previous literature that the biological brain is highly modular~\cite{caporale2008spike}, learning predominantly based on local information instead of a global objective that is optimized by backpropagating error signals~\cite{crick1989recent,marblestone2016toward}. 
In light of those observations, an intuitive approach to alleviate memory usage of MIM entails dividing the model into multiple blocks, each trained independently. 
This approach, known as~\textit{block-wise local learning}, has the potential to substantially decrease memory requirements during training since memory space can be freed up as soon as training is completed for one block. 

However, many existing local block-wise learning methods~\cite{lillicrap2014random,löwe2020putting,xiong2020loco,Belilovsky2018GreedyLL} have not been very successful in achieving performance on par with end-to-end training. This limitation is likely attributed to the fact that the learning objectives of supervised classification and contrastive learning rely heavily on global information. In contrast, MIM may require only local information to fill in missing occluded patches at different masking ratios.

Towards this end, we introduce~\textit{Block-Wise Masked Image Modeling} (BIM), which decomposes the global MIM task into several sub-tasks with independent compute patterns by implementing block-wise, rather than end-to-end, back-propagation operations. Within the BIM framework, each DNN block handles intermediate features derived from the preceding block, extracting features necessary for image reconstruction and block updates. The features that are extracted are subsequently employed to reconstruct the missing patches using a local decoder. Simultaneously, these extracted features are passed on to the next encoder block to continue their learning process. 

Moreover, to further reduce the compute workload during the training process, we introduce an incremental masking strategy to spatially increase the masking ratio without compromising learning efficacy. Experiment results indicate that our proposed method can effectively reduce peak memory while maintaining model performance without extra computation overhead.

Aside from the memory efficiency advantages offered by BIM, it seamlessly incorporates the ``Once-for-all'' training paradigm~\cite{cai2019once} through the simultaneous training of multiple ViT encoder backbones with increasing depths. This results in the generation of multiple trained DNN backbones of varying depth, each of which can adapt to different hardware platforms with unique computing capabilities. Consequently, this approach substantially diminishes computational expenses when compared to the individual training of each DNN backbone.
Overall, our contributions are summarized as follows:
\begin{itemize}
    \item We proposed BIM framework for memory and compute efficient SSL. BIM achieves a significant reduction in memory usage while delivering performance on par with traditional end-to-end training across a range of downstream tasks. Under the same setting for batch size, BIM achieves approximately an average of $40\%$ savings in peak memory and $80\%$ in compute.
    \item BIM naturally enables multiple backbone DNNs with different depth to be trained jointly, yielding a set of pre-trained backbone DNNs that can be employed independently for subsequent tasks. Compared to training each backbone DNN separately, BIM offers added computational workload savings.
    \item To further diminish the computational expenses, we introduce a novel MIM masking strategy that progressively increases the proportion of masked components over the input during BIM training, resulting in additional reductions in computational costs without losing in accuracy.

    
\end{itemize}

\section{Background and Related Work}
\label{sec:relatedwork}

\begin{figure}
\centering
\includegraphics[width=\linewidth]{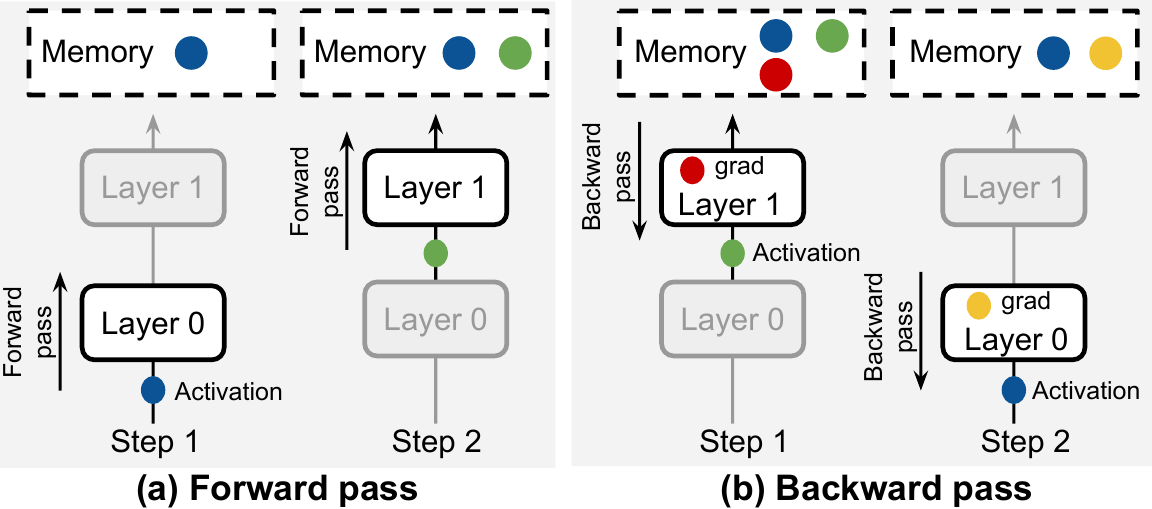}
\caption{Memory usage during forward and backward passes of DNN training.}
\label{figure:background-gpu}
\end{figure}
\subsection{Memory Pattern during DNN Training}

To provide a clearer insight into our proposed approach, we will briefly outline the memory usage pattern during the training process of the DNNs. Figure~\ref{figure:background-gpu} (a) illustrates the forward pass for a two-layer DNN. Throughout the forward pass, intermediate activations are produced once the execution of each layer is completed. The intermediate activations need be stored in memory for gradient computation during the backward pass. The same process will repeat until the DNN output is produced. The storage of intermediate activations from all layers in memory is necessary for subsequent gradient calculations, leading to an increase in memory footprint until it reaches its peak value (Step 2 in Figure~\ref{figure:background-gpu} (a)), and this peak memory usage increase in proportional with the DNN depth.

Next, the DNN output generated during the forward pass is compared to the ground-truth value in the training dataset. Subsequently, the gradient is calculated and stored in memory for use in the forthcoming backward pass, which is depicted in Figure~\ref{figure:background-gpu} (b). In the backward pass, the input activations of the last layer are initially retrieved from memory and used to compute the weight gradient by multiplication with the gradient. Additionally, the gradient is multiplied with the layer weights to generate the gradient for the preceding layers. Subsequently, the input activations can be removed, freeing up the corresponding memory space. This iterative process persists until gradients are computed for all the layer weights.

\subsection{Self-supervised Learning}
SSL has recently gained significant attention in the field of computer vision due to its outstanding accuracy performance and the fact that it does not necessitate labeled data for training. At the heart of these methods lies the formulation of a suitable pretext task that leverages the inherent information contained within the images~\cite{Doersch_2015_ICCV,oord2018representation,wang2015unsupervised}.

Among these techniques, the two most prominent approaches are contrastive learning (CL) and masked imaging model (MIM). Building upon instance discrimination~\cite{WuCvpr2018}, contrastive learning~\cite{Chen2020Simclr,Chen_2021_CVPR} strives to extract invariant latent features from images, relying heavily on effective data augmentation techniques~\cite{Grillnips2020,jeong2021contrad,wang2022contrastive,purushwalkam2020demystifying}. In contrast, MIM follows a distinct learning objective, which will be described in detailed next.

\paragraph{Masked Image Modeling.} Inspired by the great success of MLM~\cite{ben-zaken-etal-2022-bitfit,pmlr-v119-bao20a,sinha-etal-2021-masked} in NLP, researchers have explored a similar masking methodology for computer vision tasks. The primary goal of MIM is to extract information from unmasked image patches in order to reconstruct the masked patches, performing MIM can significantly enhance the feature extraction capability of backbone DNN~\cite{xie2021simmim}. Existing research in the field has shown a preference for using large batch sizes (e.g., 4096), to train the backbone DNN with large number of epochs (e.g., 800), and has empirically observed improvements in MIM performance as a result. However, this places significant demands on memory capacity and compute capability, limiting the application of MIM. While some previous research has introduced advanced masking strategies to reduce the computation workload \cite{kakogeorgiou2022attmask,li2022semmae,liu2022good,Kong_2023_CVPR,Wang_2023_CVPR}, hardly any of these approaches have addressed the reduction of peak memory size. In this study, we employ the masked autoencoder (MAE)~\cite{He2022MAE}, a pioneering framework for masked image modeling, as a benchmark to demonstrate how our method effectively reduces peak memory consumption and computational cost.

\paragraph{Local Learning Paradigms.} 
Compared to end-to-end supervised learning approaches~\cite{dosovitskiy2020vit,He2016resnet,lecun1998cnn}, SSL methods~\cite{xie2021simmim,He2022MAE} require a considerably higher memory requirement due to larger batch size.
Local learning has emerged as a practical strategy for addressing the issue of high memory demand. It achieves substantial memory reduction by segmenting the model into separate gradient-isolated blocks and enabling their independent training. As a pioneering method for gradient-isolated block-wise training, the Greedy InfoMax algorithm~\cite{Belilovsky2018GreedyLL} shows impressive performance on small datasets, using only $40\%$ of the memory compared to what traditional supervised learning methods typically demand. However, the performance of InfoMax will degrade seriously on large-scale datasets, such as ImageNet. LoCo~\cite{xiong2020loco} is a local learning algorithm for unsupervised contrastive learning. By leveraging implicit gradient feedback between the gradient-isolation blocks, LoCo can achieve results comparable to those of the conventional contrastive learning framework. Nevertheless, LoCo comes with a significantly higher computational cost compared to the end-to-end training approach. In comparison, BIM enables a MIM-based SSL pretraining with substantial reductions in peak memory consumption and computational workload.

\paragraph{Efficient DNN Training.} 
\looseness=-10000
Previous research has explored various techniques to expedite DNN training through leveraging sparsity in weights and activations~\cite{mahmoud2020tensordash, zhang2019eager, yang2020procrustes, choi2020energy, qin2020sigma, zhang2022fast}. For instance, Procrustes~\cite{yang2020procrustes} and Eager Pruning~\cite{zhang2019eager} improve training efficiency by aligning algorithms with hardware capabilities, eliminating unimportant DNN weights, and enhancing hardware efficiency. Another approach involves reducing DNN operand precision~\cite{judd2016stripes, lee20197} or dynamically adjusting precision during DNN training, as proposed in FAST~\cite{zhang2022fast}. These techniques are orthogonal to our BIM framework and can provide additional training efficiency. There are also works focused on reducing memory consumption during DNN training by modifying the DNN architecture~\cite{gomez2017reversible,zhang2023camel}. In comparison, BIM represents a versatile training framework applicable to all DNN architectures.
\section{Method}
\label{sec:method}

\begin{figure*}
\centering
\includegraphics[width=0.95\linewidth]{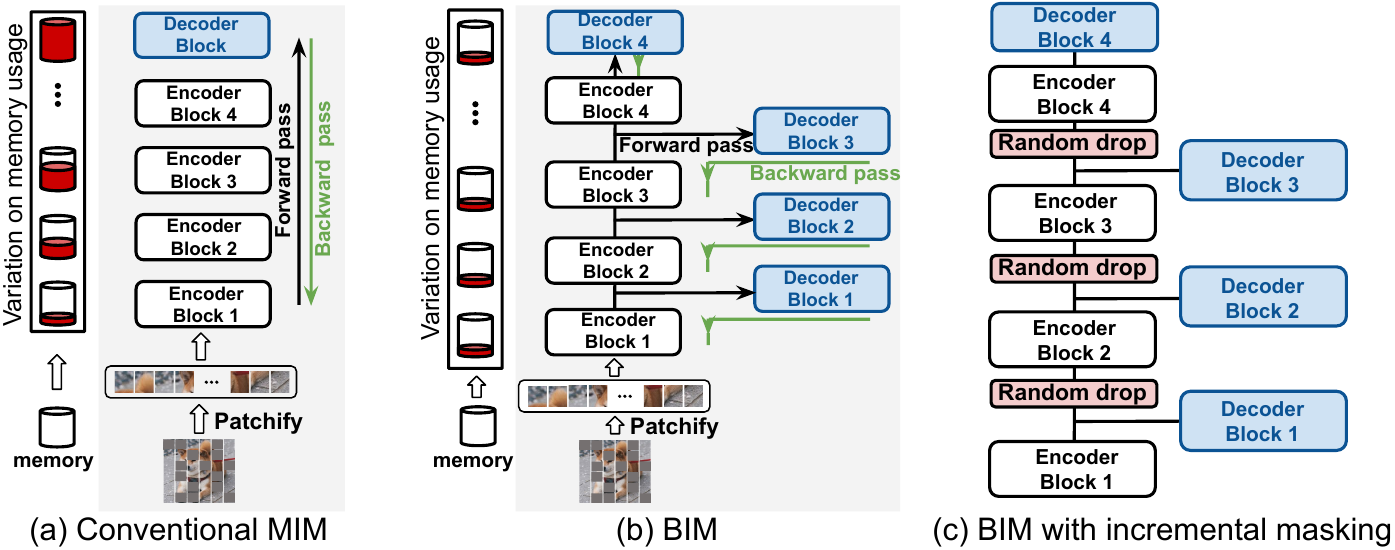}
\caption{(a) Training flow for Conventional MIM. (b) Training flow for BIM. (c) Overview of incremental masking strategy. }
\label{figure:framework}
\end{figure*}

In this section, we first present our memory-efficient BIM paradigm in Section~\ref{method-overview}. Next, Section~\ref{sec:once-for-all} describes the ``Once-for-all'' training paradigm offered by BIM, which enables the jointly training over multiple backbone DNNs. In Section~\ref{method-increasing-mask-ratio}, we elaborate on the incremental masking strategy, which serves to minimize the computational overhead resulting from additional decoders. Finally, in Section~\ref{method-GPU-scheduling}, we describe the memory scheduling for BIM, explaining how it significantly reduces the strain on memory capacity.

\subsection{DNN Pre-training with BIM}
\label{method-overview}
The conventional MIM training process is illustrated in Figure~\ref{figure:framework} (a), where an end-to-end training algorithm is applied to all encoder and decoder blocks of ViT. This necessitates the allocation of extensive memory to store the entire set of model weights and gradients, resulting in a substantial demand for memory. To mitigate this, we proposed a block-wise local learning approach. An illustrative representation of our novel framework is presented in Figure \ref{figure:framework} (b). Before training, the stack of ViT encoders are initially divided into several blocks of uniform size, each associated with a decoder of the corresponding size. Subsequently, each ViT encoder-decoder pair undergoes separate training with the same objective loss function. 

Specifically, the computational flow during the forward pass in BIM closely resembles that of MIM. In this process, each ViT encoder block receives intermediate results from the preceding blocks, conducts forward computations, and generates an output. This output is then duplicated: one copy is forwarded to the next encoder block for further processing, while the other copy is sent to the corresponding decoder block. The decoder's role is to produce predictions for masked patches using the features extracted by the current encoder block. This procedure continues until all the decoders have generated their predictions. The reconstructed patches from each decoder are subsequently compared to the original unmasked image patches, resulting in the generation of gradients for the backward pass operations.


With the gradients generated from the loss function at each decoder, the backward pass operations are carried out independently within each ViT encoder-decoder block. To be specific, the computation of gradients is terminated when it reaches the beginning of the current ViT encoder block (Figure~\ref{figure:framework} (b)). This ensures that there is no overlap in gradients across different blocks, and the gradient from one ViT encoder block does not influence the earlier ViT encoders. As a result of this gradient isolation strategy, in conjunction with our memory scheduling algorithm detailed in Section~\ref{method-GPU-scheduling}, BIM results in a significant reduction in peak memory usage than the conventional MIM.

\subsection{Once-for-all Pre-training with BIM}
\label{sec:once-for-all}

In addition to the memory efficiency benefits provided by BIM, it also naturally implements the ``Once-for-all'' training paradigm~\cite{cai2019once} by joint training of multiple ViT encoder backbones with growing depths. To illustrate this, if a ViT encoder is divided into four blocks, BIM allows the training of four distinct backbone DNNs with increasing depths by truncating at the output of each encoder block. This ``Once-for-all'' paradigm offered by BIM empowers the resultant pre-trained model to adapt to different computational constraints and diverse tasks, optimizing resource utilization and versatility. Compared to the conventional approach of separately training each backbone DNN with different depths, BIM results in substantial computational savings.

\subsection{Incremental Masking Ratio Growth with BIM}
\label{method-increasing-mask-ratio}
We also introduce an incremental masking strategy that progressively increases the proportion of masked patches, aiming to achieve additional computational savings.
Specifically, we enhance the level of difficulty in BIM by progressively decreasing the proportion of unmasked patches used for image reconstruction. As shown in Figure~\ref{figure:framework} (c), we introduce a new layer at the end of each encoder block, which randomly discards additional patches during the forward computation. The percentage of these additional drops is predefined and increases with the layer depth. This approach results in a reduction in the input size for each encoder block, effectively reducing the computational workload while obtaining a comparable (even better) performance as described in Section~\ref{sec:ablations}.


\subsection{BIM Computation Pattern}
\label{method-GPU-scheduling}
\begin{figure*}
\centering
\includegraphics[width=0.95\linewidth]{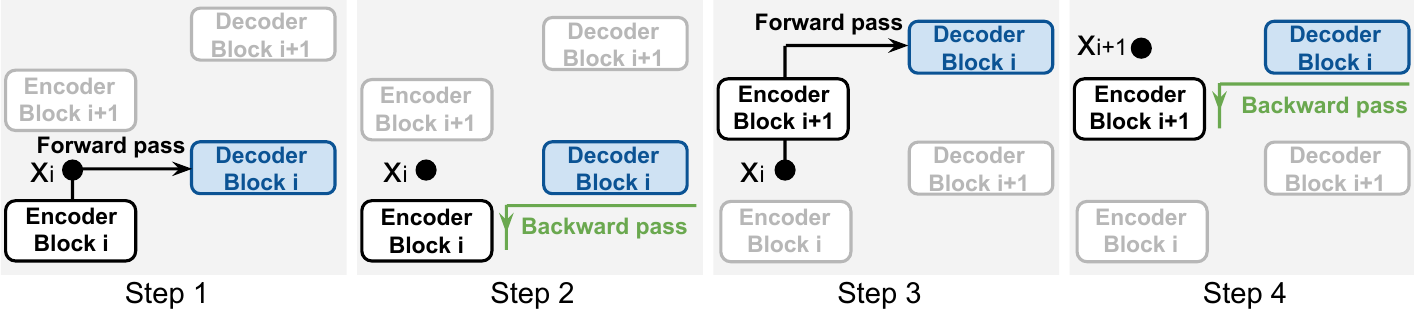}
\caption{The computation pattern of BIM over two consecutive encoder blocks. The block dot denotes the intermediate activation during the forward pass.}
\label{figure:GPU-scheduling}
\end{figure*}
We describe an efficient computation pattern for both forward and backward passes of BIM training. The proposed computation pattern can achieve optimal peak memory consumption. We illustrate this scheduling algorithm with an example that show the computation pattern within the two consecutive encoder blocks (Figure~\ref{figure:GPU-scheduling}).

Initially, the input patches $x_{i-1}$ are fed into the encoder block $\textbf{i}$ to generate output $x_{i}$. All the intermediate activations, including $x_{i}$, are buffered in the memory for later use. (Step 1 in Figure~\ref{figure:GPU-scheduling}). $x_{i}$ is subsequently input into decoder block $\textbf{i}$, resulting in the generation of predicted outputs and the initiation of the loss gradient. The backward pass starts as indicated in Step 2, results in additional weight updates in both decoder $i$ and encoder $i$. This backward pass concludes when it reaches the initial layer of encoder $i$. Once the parameter updates in encoder block $\textbf{i}$ and decoder block $\textbf{i}$ are finished, all intermediate features stored in the buffer, except for $x_{i}$, can be cleared from memory, preserving them for future use. $x_{i}$ is subsequently forwarded to encoder block $\textbf{i+1}$, and the identical process repeats, as illustrated in Steps 3 and 4 in Figure~\ref{figure:GPU-scheduling}.

\section{Experiments}
\label{sec:Experiment}
In this section, we conduct experiments to validate the BIM performance. We describe the implementation details in Section~\ref{sec:exp-implementation} and presents the main results in Section~\ref{sec:results}. The ablation studies are shown in Section~\ref{sec:ablations}. 

\subsection{Implementation}
\label{sec:exp-implementation}
\textbf{Pretrain model on ImageNet-1K}
We evaluate BIM by comparing its performance against MAE~\cite{He2022MAE}, and reports its performance in terms of accuracy and peak memory consumption. We compare BIM with MAE over different versions of ViT, including ViT-base, ViT-large, and ViT-huge. The models are pretrained for either 400 or 800 epochs on the ImageNet-1k dataset \cite{DengImagenet}. For BIM training of ViT-base and ViT-large, we divide their encoder backbone into four blocks and train it with a batch size of 4096 or 8192, whereas for ViT-huge, we partition it into two blocks and train it with a batch size of 2048 or 4096. We use a base learning rate of $1.5\times 10^{-4}$ with a cosine decay schedule without restart. Linear warm-up lasts the first 40 epochs. 

All the pretraining computations are executed on a GPU cluster consisting of 4 nodes, each node equipped with 8 NVIDIA V100 GPUs. We adopt a fixed masking ratio with a ratio of $75\%$, and we will test incremental masking ratio in Section~\ref{sec:ablations}. We follow the original MAE work~\cite{He2022MAE} for the rest of the training settings.
Pretrained ViT encoder backbones are evaluated over two downstream tasks, which are elaborated upon as follows.

\paragraph{Image Classification.}
We utilize pretrained ViT encoder backbones as the initializations. Subsequently, we append a linear layer to the pretrained encoder backbone and perform either end-to-end fine-tuning or linear probing on the ImageNet-1K dataset. In the case of end-to-end fine-tuning, the entire model undergoes fine-tuning, whereas for linear probing, only the weights of the linear layer are modified.


\paragraph{COCO Detection and Instance Segmentation.}
Additionally, we assess the performance of the pretrained backbone with tasks of object detection and instance segmentation. More precisely, we take Mask R-CNN~\cite{he2017mask} as the objective detector and perform end-to-end finetuning on the COCO dataset~\cite{lin2015microsoft}, among which the ViT backbone is adapted for use with FPN~\cite{lin2017feature}, following ViTDet framework~\cite{li2022exploring}. The size of the input image is $1024\times 1024$, augmented with large-scale jittering~\cite{ghiasi2021simple} during training. We fine-tune both the ViT-base and ViT-large backbone 100 epochs with the AdamW optimizer. Notably, the feature map scale for both backbone architectures is set to 1/16, \ie, stride=16, since the patch size is 16. We apply this approach to all entries in Table~\ref{tab:overall-result}. We report box AP for object detection and mask AP for instance segmentation. 


\subsection{Main Results}
\label{sec:results}

\begin{figure*}
\centering
\includegraphics[width=\linewidth]{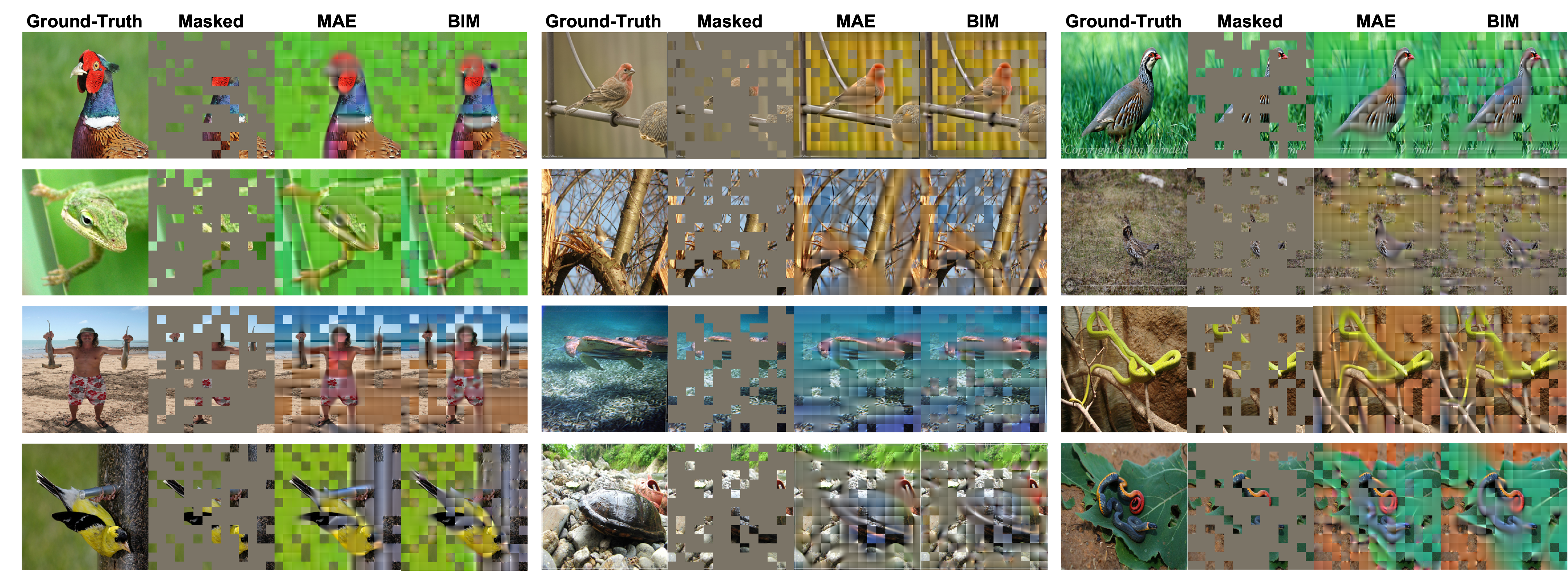}
\caption{Results on image reconstruction. For each set of four images, we display them in the following order from left to right: ground-truth, masked image, MAE reconstruction, and our BIM reconstruction. 
The backbone DNN is extracted from ViT-large model pretrained with either BIM or MAE with 800 epochs. The weights for the MAE backbone is downloaded from the official website~\cite{mae_repo}.}
\label{figure:reconstruction-result}
\end{figure*}

\paragraph{Masked Image Reconstruction.}
\looseness=-10000
To evaluate the masked image reconstruction performance, we visualize some sample images that are taken from the validation set of ImageNet. As Figure~\ref{figure:reconstruction-result} shows, for each quadruple, from left to right, we present the ground-truth images, the images with $75\%$ masking, the MAE reconstruction results, and BIM reconstruction results, respectively. 
The reconstruction visualization demonstrates that the ViT backbones trained with BIM can achieve a strong image reconstruction capability.

\paragraph{End-to-End Fine-tuning.}
\begin{table*}[t]
\centering
\resizebox{\linewidth}{!}{
\begin{tabular}{c|c|c|c|c|c|c|c|c}
\bottomrule[1pt]
Backbone&\makecell{Patch\\Size}&\makecell{Pretrain\\Epoch}&\makecell{Pretrain\\Batch Size}& \makecell{Pretrain\\Method}&\makecell{Num of \\ Blocks}&\makecell{Peak Memory \\ Consumption (GB)}&\makecell{Fine-tuning \\ Top-1 acc ($\%$)}&\makecell{Linear Eval\\ Top-1 acc ($\%$)}\\

\bottomrule[1pt]
\multirow{6}{*}{ViT-base}&\multirow{6}{*}{$16^{2}$}&\multirow{3}{*}{400}&\multirow{2}{*}{4096}&MAE&-&1218.57 ($1\times$)&83.01&63.03\\
\cline{5-9}
&&&&BIM&4&929.09 (\textcolor{green}{$0.76\times$})&82.98 \textcolor{green}{$\downarrow 0.03$}&62.87\textcolor{green}{$\downarrow 0.16$}\\
\cline{4-9}
&&&8192&BIM&4&1857.99 (\textcolor{red}{$1.52\times$})&83.56\textcolor{red}{$\uparrow 0.55$}&65.46\textcolor{red}{$\uparrow 2.43$}\\
\cline{3-9}
&&\multirow{3}{*}{800}&\multirow{2}{*}{4096}&MAE&-&1218.57 $(1\times)$&83.27&66.25\\
\cline{5-9}
&&&&BIM&4&929.09 (\textcolor{green}{$0.76\times$})&83.20\textcolor{green}{$\downarrow 0.07$}&66.08\textcolor{green}{$\downarrow 0.17$}\\
\cline{4-9}
&&&8192&BIM&4&1857.99 (\textcolor{red}{$1.52\times$})&83.89\textcolor{red}{$\uparrow 0.62$}&69.24\textcolor{red}{$\uparrow 2.99$}\\

\bottomrule[1pt]
\multirow{6}{*}{ViT-large}&\multirow{6}{*}{$16^{2}$}&\multirow{3}{*}{400}&\multirow{2}{*}{4096}&MAE&-&1865.58 ($1\times$)&84.79&70.20\\
\cline{5-9}
&&&&BIM&4&1093.51 (\textcolor{green}{$0.59\times$})&84.58\textcolor{green}{$\downarrow 0.21$}&68.60\textcolor{green}{$\downarrow 1.60$}\\
\cline{4-9}
&&&8192&BIM&4&2186.64 (\textcolor{red}{$1.17\times$})&85.37\textcolor{red}{$\uparrow 0.58$}&74.07\textcolor{red}{$\uparrow 1.87$}\\
\cline{3-9}
&&\multirow{3}{*}{800}&\multirow{2}{*}{4096}&MAE&-&1865.58 ($1\times$)&85.15&73.93\\
\cline{5-9}
&&&&BIM&4&1093.51 (\textcolor{green}{$0.59\times$})&84.99\textcolor{green}{$\downarrow 0.16$}&73.10\textcolor{green}{$\downarrow 0.83$}\\
\cline{4-9}
&&&8192&BIM&4&2186.64 (\textcolor{red}{$1.17\times$})&85.67\textcolor{red}{$\uparrow 0.52$}&76.51\textcolor{red}{$\uparrow 3.41$}\\

\bottomrule[1pt]
\multirow{6}{*}{ViT-huge}&\multirow{6}{*}{$14^{2}$}&\multirow{3}{*}{400}&\multirow{2}{*}{2048}&MAE&-&1695.30 ($1\times$)&86.12&73.62\\
\cline{5-9}
&&&&BIM&2&1136.40 (\textcolor{green}{$0.67\times$})&86.03\textcolor{green}{$\downarrow 0.09$}&72.54\textcolor{green}{$\downarrow 1.08$}\\
\cline{4-9}
&&&4096&BIM&2&2271.50 (\textcolor{red}{$1.34\times$})&86.41\textcolor{red}{$\uparrow 0.29$}&74.75\textcolor{red}{$\uparrow 2.21$}\\
\cline{3-9}
&&\multirow{3}{*}{800}&\multirow{2}{*}{2048}&MAE&-&1695.30 ($1\times$)&86.38&76.77\\
\cline{5-9}
&&&&BIM&2&1136.40 (\textcolor{green}{$0.67\times$})&86.24\textcolor{green}{$\downarrow 0.14$}&76.23\textcolor{green}{$\downarrow 0.54$}\\
\cline{4-9}
&&&4096&BIM&2&2271.50 (\textcolor{red}{$1.34\times$})&86.60\textcolor{red}{$\uparrow 0.22$}&77.64\textcolor{red}{$\uparrow 0.87$}\\
\bottomrule[1pt]
\end{tabular}
}
\caption{Accuracy comparison for ViT backbones pretrained with BIM and MAE over Image classification task. BIM achieves a comparable accuracy performance as MAE while greatly saves the peak memory.}
\label{tab:overall-result}
\end{table*}
For the end-to-end fine-tuning approach, we initiate the ViT by utilizing the pretrained encoder backbone and subsequently fine-tune the entire network.
Specifically, we fine-tune ViT-base for 100 epochs and ViT-large and ViT-huge for 50 epochs, adhering to the procedures detailed in~\cite{He2022MAE}. We assess the performance of our BIM with MAE in terms of accuracy and peak GPU memory consumption across multiple ViT backbones. The results are presented in Table~\ref{tab:overall-result}, revealing some noteworthy findings. Firstly, under identical settings for training epochs and batch size, BIM consistently achieves comparable accuracies, with an average difference of less than $0.1\%$, in the end-to-end fine-tuning task across different backbone architectures. Secondly, when pretrained with a batch size of 8192, BIM attains an average fine-tuning accuracy that is 0.47 higher than that MAE that are pretrained with a batch size of 4096, across various DNN backbones.




\paragraph{Linear Probing.}
In addition to the end-to-end fine-tuning, we also assess the linear probing scheme, where only the linear layer undergoes fine-tuning. As shown in Table~\ref{tab:overall-result}, our proposed BIM demonstrates superior performance, exhibiting only an average accuracy drop of $0.52\%$ when compared to the traditional MAE approach.


\paragraph{Transfer Learning on COCO.}
\begin{table}[t]
\centering
\resizebox{\columnwidth}{!}{%
\begin{tabular}{|c|c|c|c|c|c|}
\hline 
\multirow{2}{*}{\makecell{Pretrain\\Method}}&\multirow{2}{*}{\makecell{Pretrain\\Dataset}}&\multicolumn{2}{|c|}{ViT-base}&\multicolumn{2}{|c|}{ViT-large}\\
\cline{3-6}
& & $AP^{box}$ & $AP^{mask}$ & $AP^{box}$ & $AP^{mask}$ \\
\hline
MAE&\multirow{2}{*}{ImageNet}& 51.2& 45.5 & 54.6 & 48.6\\
\cline{1-1}\cline{3-6}
BIM&& 51.2& 45.3 & 54.3 & 48.2\\
\hline
\end{tabular}
}%
\caption{Transfer learning results on \textbf{COCO} object detection and segmentation using \textbf{ViTDet}.}
\label{tab:transfer-learning}
\end{table}

To assess the transferability of the features derived from our proposed framework, we conduct end-to-end fine-tuning of Mask R-CNN for ViT on the COCO dataset using pretrained backbone weights from ViT-base and ViT-large, each pretrained for 800 epochs. Following the ViTDet implementation~\cite{li2022exploring}, the ViT backbone is integrated with the FPN, applied uniformly across all pre-trained entities. The results, as shown in Table~\ref{tab:transfer-learning}, demonstrate that ViT backbones pretrained with BIM can also effectively transfer to object detection and instance segmentation tasks, yielding performance comparable to MAE.


\paragraph{Saving on Peak Memory.}

\begin{figure*}
\centering
\includegraphics[width=\linewidth]{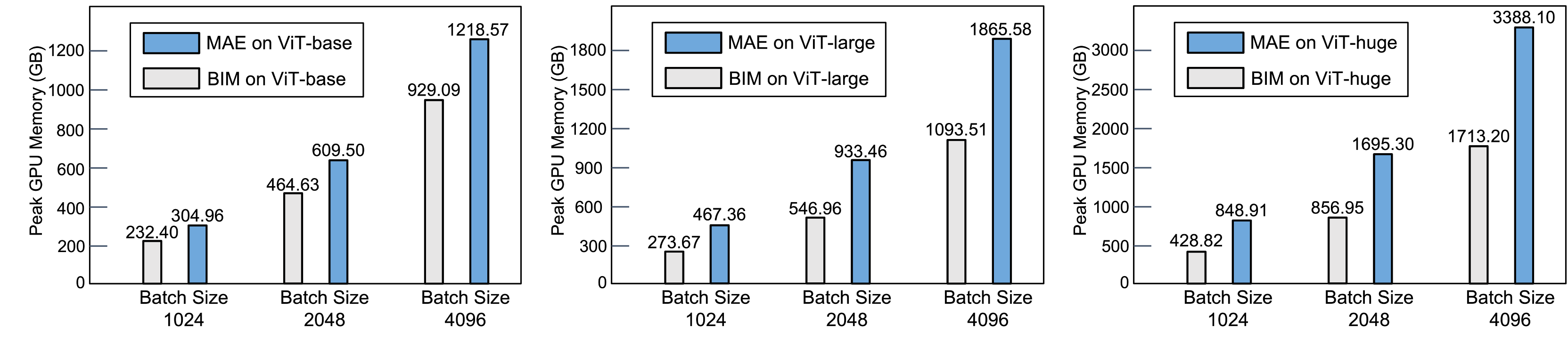}
\caption{Peak memory usage comparison between BIM and MAE with varying ViT backbones and batch sizes.}
\label{figure:GPU-consumption}
\end{figure*}
One of the key advantages offered by BIM is the substantial reduction in peak GPU memory usage. To visualize this, Figure~\ref{figure:GPU-consumption} provide a peak memory consumption comparison between MAE and BIM across various batch size, under the settings that all the ViT backbones are divided into four blocks for BIM training. We notice that BIM achieves an average of $25\%$, $42\%$ and $48\%$ peak memory savings than MAE on ViT-base, ViT-large and ViT-huge, respectively. Specifically, the peak memory savings increase as the ViT encoder backbone size increases. This is due to the diminishing impact of the ViT decoder on memory usage as the encoder size grows. Ignoring the impact of other components such as decoders and embedding layers, BIM training of a ViT backbone with four blocks using BIM can result in up to a $4\times$ reduction in peak memory usage. 

BIM offers a promising solution for researchers facing constraints in GPU resources. Specifically, BIM enables a ViT backbone to be trained with a larger batch size, and therefore leading to better accuracies over the downstream tasks.




\paragraph{Performance under the Same Peak Memory.}
In this section, we evaluate the performance of BIM and MAE while maintaining a fixed peak memory consumption. To achieve this, we adjust the batch size during BIM training on ViT-large so that its peak memory usage matches that of MAE with a batch size of 4096. This results in a batch size of 6784 for both ViT-large. We then train ViT-large with BIM for 400 epochs using batch size of 6784 and finetune it over the ImageNet dataset with 50 epochs. This leads to an accuracy of $85.02\%$, which is better than that obtained with MAE backbone ($84.79\%$). 

\paragraph{Once-for-all Training of BIM.}
\begin{table}[h]
\centering
\resizebox{\columnwidth}{!}{%
\begin{tabular}{|c|c|c|c|c|c|c|}
\hline 
\multirow{2}{*}{\makecell{Pretrain\\Method}}&\multirow{2}{*}{Backbone}&\multicolumn{4}{|c|}{Acc with Num of Blocks (\%)}&\multirow{2}{*}{\makecell{Training\\Cost Saving (\%)}}\\
\cline{3-6}
&&1&2&3&4&\\
\hline
IT&\multirow{3}{*}{ViT-base}&70.59&79.04&81.48&83.01&-\\
\cline{1-1}\cline{3-7}
MD&&65.88&75.02&79.46&83.01&61.12\\
\cline{1-1}\cline{3-7}
BIM&&70.59&79.06&81.39&82.98&\textbf{68.00}\\
\hline
IT&\multirow{3}{*}{ViT-large}&77.78&82.49&84.22&84.79&-\\
\cline{1-1}\cline{3-7}
MD&&73.70&79.41&82.78&84.79&78.23\\
\cline{1-1}\cline{3-7}
BIM&&77.78&82.3&84.14&84.58&\textbf{85.18}\\
\hline
IT&\multirow{3}{*}{ViT-huge}&80.75&84.88&85.53&86.12&-\\
\cline{1-1}\cline{3-7}
MD&&75.78&80.34&83.48&86.12&80.85\\
\cline{1-1}\cline{3-7}
BIM&&80.75&84.52&85.50&86.03&\textbf{87.84}\\
\hline
\end{tabular}
}%
\caption{Fine-tuning performance comparison among IT, MD, and BIM. BIM performs much better than MD, while achieving a comparable performance as IT. Compared with IT, BIM significantly reduces training cost by up to $87.84\%$.}
\label{tab:sub-vit}
\end{table}
As discussed in Section~\ref{sec:once-for-all}, BIM offers a natural advantage by allowing the simultaneous training of multiple ViT-encoder backbones with varying depths, all sharing their weights in a nested manner. Our evaluation of ViT encoders' performance is conducted on the ImageNet dataset. Specifically, we compare our approach with two alternative baseline methods. The first baseline, referred to as ~\textit{independent training (IT)}, involves training each ViT encoder backbone with varying depth separately. Note that this approach incurs significantly higher training costs since each ViT encoder is trained independently. The second baseline, denoted as~\textit{MAE directly (MD)}, entails truncating the pretrained MAE at the end of each ViT encoder block and fine-tuning it for downstream tasks. All ViT backbones are pretrained for 400 epochs, then fine-tuned with 100 epochs for ViT-base and 50 epochs for ViT-large and ViT-huge.


As indicated in Table \ref{tab:sub-vit}, we note that BIM outperforms MD by a significant margin and closely matches the performance of IT. In particular, for ViT with a single encoder block, BIM and IT exhibit identical training schemes, resulting in the same accuracy. Notably, BIM achieves an average of $77\%$ reduction in training cost compared to IT. 


\subsection{Ablation Study}
\label{sec:ablations}
\begin{table}[t]
\centering
\resizebox{\columnwidth}{!}{%
\begin{tabular}{|c|c|c|c|c|c|c|}
\hline 
\multirow{2}{*}{Method}&\multicolumn{5}{|c|}{Mask Ratio (\%)}&\multirow{2}{*}{\makecell{Fine-tune Top-1 \\ Acc (\%)}}\\
\cline{2-6}
&Block 1&Block 2&Block 3&Block 4&On Average&\\
\hline
\multirow{5}{*}{\makecell{Fixed\\Ratio}}&\multicolumn{5}{|c|}{60}&81.79\\
\cline{2-7}
&\multicolumn{5}{|c|}{70}&83.80\\
\cline{2-7}
&\multicolumn{5}{|c|}{75}&84.58\\
\cline{2-7}
&\multicolumn{5}{|c|}{80}&84.28\\
\cline{2-7}
&\multicolumn{5}{|c|}{90}&82.67\\
\hline
\multirow{2}{*}{\makecell{Incremental \\ Masking Ratio}}&75&80&85&90&82.5& 84.44\\
\cline{2-7}
&65&70&80&85&75&\textbf{84.75}\\
\hline
\end{tabular}
}%
\caption{Comparison between fixed masking ratio and incremental masking ratio.}
\label{tab:ablation-masking}
\end{table}

\paragraph{Impact of Masking Ratio.}
We evaluate the performance of BIM over different masking ratio. Specifically, we want to investigate two problems. Firstly, we examine how BIM accuracy is influenced by different masking ratios, assuming all the blocks have the identical masking ratio. Secondly, we investigate how accuracy varies when adjusting the masking ratio across different ViT encoder blocks. 

We trained with the ViT-large encoder backbone using various masking ratios for 400 epochs and subsequently fine-tuned on the ImageNet dataset for 50 epochs. As shown in Table~\ref{tab:ablation-masking}, we observe that under the assumption that all the blocks have the same masking ratio, a masking ratio of $75\%$ yields the best overall performance, which aligns with findings from the MAE approach~\cite{He2022MAE}. Furthermore, it is worth noting that incremental masking ratios in general outperforms constant masking ratios under the same average masking ratio. In particular, we find that masking ratios of $65\%, 70\%, 80\%, 85\%$, with an average masking ratio of $75\%$, yield superior performance compared to a fixed masking ratio of $75\%$ across all blocks. In comparison, employing a combination of masking ratios such as $75\%$, $80\%$, $85\%$, and $90\%$ with an average masking ratio of $82.5\%$ results in an accuracy of $84.65\%$. This performance is comparable to that achieved with an average masking ratio of $75\%$, while greatly reducing computational workload.



\paragraph{Impact of Block Numbers.}
\begin{figure}
\centering
\includegraphics[width=1.2\linewidth]{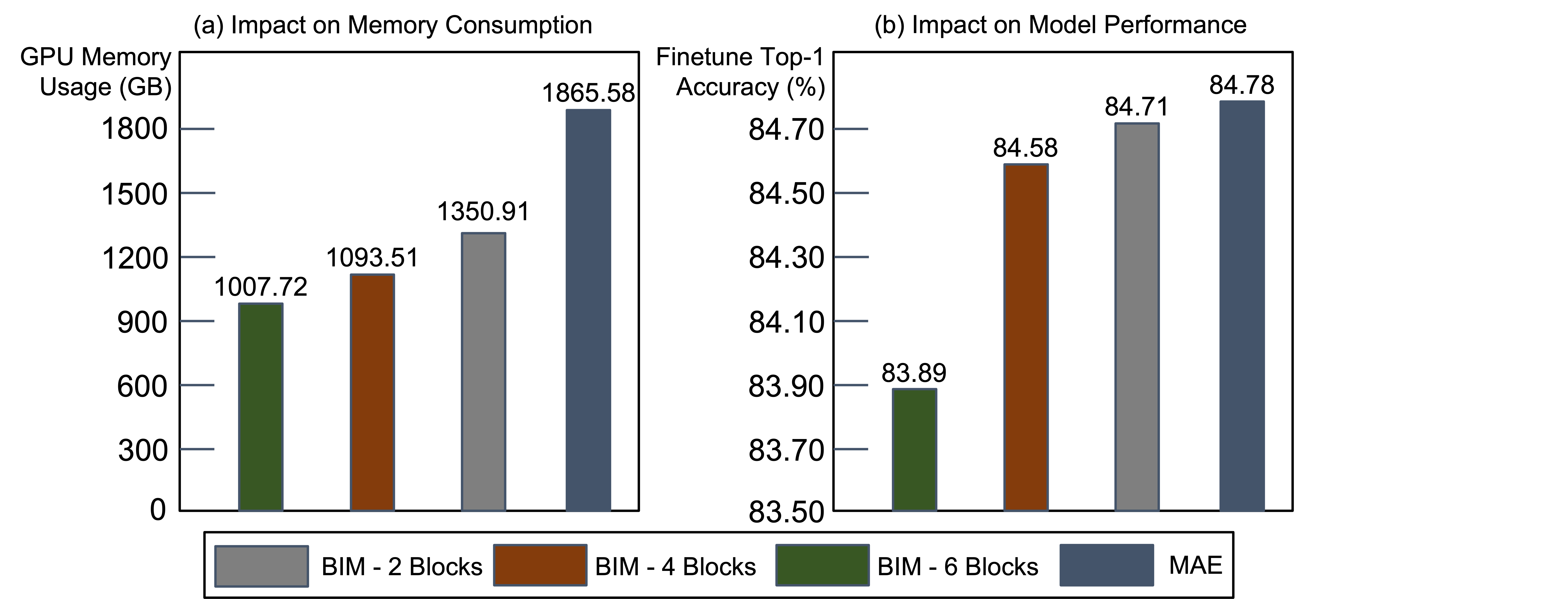}
\caption{Ablation study on the number of ViT blocks. (a) Impact on Peak memory consumption. (b) Impact on the accuracy performance over the ImageNet. }
\label{figure:ablation-num-of-blocks}
\end{figure}

In this section, we explore how the number of ViT blocks affects the accuracy of BIM. We specifically investigate how varying the number of blocks impacts BIM's performance. To achieve this, we pretrained the ViT-large backbone with 400 epochs using 2 blocks, 4 blocks, and 6 blocks, followed by fine-tuning the entire ViT-large model on ImageNet with 50 epochs.

The findings presented in Figure~\ref{figure:ablation-num-of-blocks} reveal that a larger number of blocks indeed results in lower peak memory usage for BIM. However, this reduction in memory consumption comes at the cost of decreased model performance. For example, when transitioning from 4 to 6 blocks, BIM achieves only a marginal $4\%$ reduction in memory consumption while incurring a notable $0.7\%$ drop in accuracy, which may not be justified.




\paragraph{Impact on Training Iteration.}

\begin{table}[t]
\centering
\resizebox{\columnwidth}{!}{%
\begin{tabular}{c|c|c|c|c}
\hline 
Iterations & 400 epochs & 800 epochs & 1200 epochs & 1600 epochs \\
\hline
BIM Acc. (\%) & 84.58 & 85.09 & 85.15 & 85.27\\
MAE Acc. (\%) & 84.79 & 85.15 & 85.23 & 85.38\\
\hline
\end{tabular}
}%
\caption{Experiment on increasing pretraining epochs. Both MAE and BIM benefit from longer pretraining epochs.}
\label{tab:impact-iteration}
\end{table}

Finally, we investigate the effect of training iteration on accuracy during the fine-tuning phase for downstream tasks. To this end, we train ViT-large using BIM and MAE for varying numbers of epochs: 400, 800, 1200, and 1600, followed by fine-tuning on the ImageNet dataset for 50 epochs. The results, as shown in Table~\ref{tab:impact-iteration}, indicate a general trend of increasing accuracy with training iteration for both BIM and MAE. However, it is worth noting that the incremental gain in accuracy is relatively modest when extending training from 800 epochs to 1600 epochs.


\label{subsec:ablation-study}

\section{Conclusion}

Compared to other training methods like supervised learning and unsupervised contrastive learning, MIM pretraining typically demands significant computational resources, especially when handling large training data batches. The substantial memory requirements and computational cost associated with MIM pose a significant challenge to its widespread adoption. 

In this work, we introduce BIM as an alternative solution for SSL pretraining, achieving comparable performance to MIM while significantly reducing peak memory consumption. This opens up interesting future avenues in a promising direction of research.

{
    \newpage
    \small
    \bibliographystyle{ieeenat_fullname}
    \bibliography{main}

\begin{thebibliography}{48}
\providecommand{\natexlab}[1]{#1}
\providecommand{\url}[1]{\texttt{#1}}
\expandafter\ifx\csname urlstyle\endcsname\relax
  \providecommand{\doi}[1]{doi: #1}\else
  \providecommand{\doi}{doi: \begingroup \urlstyle{rm}\Url}\fi

\bibitem[mae(2021)]{mae_repo}
Masked autoencoder official implementation.
\newblock \emph{https://github.com/facebookresearch/mae/}, 2021.

\bibitem[Bao et~al.(2020)Bao, Dong, Wei, Wang, Yang, Liu, Wang, Gao, Piao, Zhou, and Hon]{pmlr-v119-bao20a}
Hangbo Bao, Li Dong, Furu Wei, Wenhui Wang, Nan Yang, Xiaodong Liu, Yu Wang, Jianfeng Gao, Songhao Piao, Ming Zhou, and Hsiao-Wuen Hon.
\newblock {U}ni{LM}v2: Pseudo-masked language models for unified language model pre-training.
\newblock In \emph{Proceedings of the 37th International Conference on Machine Learning}, pages 642--652. PMLR, 2020.

\bibitem[Belilovsky et~al.(2018)Belilovsky, Eickenberg, and Oyallon]{Belilovsky2018GreedyLL}
Eugene Belilovsky, Michael Eickenberg, and Edouard Oyallon.
\newblock Greedy layerwise learning can scale to imagenet.
\newblock In \emph{International Conference on Machine Learning}, 2018.

\bibitem[Ben~Zaken et~al.(2022)Ben~Zaken, Goldberg, and Ravfogel]{ben-zaken-etal-2022-bitfit}
Elad Ben~Zaken, Yoav Goldberg, and Shauli Ravfogel.
\newblock {B}it{F}it: Simple parameter-efficient fine-tuning for transformer-based masked language-models.
\newblock In \emph{Proceedings of the 60th Annual Meeting of the Association for Computational Linguistics (Volume 2: Short Papers)}, pages 1--9, Dublin, Ireland, 2022. Association for Computational Linguistics.

\bibitem[Cai et~al.(2019)Cai, Gan, Wang, Zhang, and Han]{cai2019once}
Han Cai, Chuang Gan, Tianzhe Wang, Zhekai Zhang, and Song Han.
\newblock Once-for-all: Train one network and specialize it for efficient deployment.
\newblock \emph{arXiv preprint arXiv:1908.09791}, 2019.

\bibitem[Caporale and Dan(2008)]{caporale2008spike}
Natalia Caporale and Yang Dan.
\newblock Spike timing--dependent plasticity: a hebbian learning rule.
\newblock \emph{Annu. Rev. Neurosci.}, 31:\penalty0 25--46, 2008.

\bibitem[Chen et~al.(2020)Chen, Kornblith, Norouzi, and Hinton]{Chen2020Simclr}
Ting Chen, Simon Kornblith, Mohammad Norouzi, and Geoffrey Hinton.
\newblock A simple framework for contrastive learning of visual representations.
\newblock In \emph{Proceedings of the 37th International Conference on Machine Learning}. JMLR.org, 2020.

\bibitem[Chen and He(2021)]{Chen_2021_CVPR}
Xinlei Chen and Kaiming He.
\newblock Exploring simple siamese representation learning.
\newblock In \emph{Proceedings of the IEEE/CVF Conference on Computer Vision and Pattern Recognition (CVPR)}, pages 15750--15758, 2021.

\bibitem[Choi et~al.(2020)Choi, Sim, Kang, Choi, Kim, and Kim]{choi2020energy}
Seungkyu Choi, Jaehyeong Sim, Myeonggu Kang, Yeongjae Choi, Hyeonuk Kim, and Lee-Sup Kim.
\newblock An energy-efficient deep convolutional neural network training accelerator for in situ personalization on smart devices.
\newblock \emph{IEEE Journal of Solid-State Circuits}, 55\penalty0 (10):\penalty0 2691--2702, 2020.

\bibitem[Crick(1989)]{crick1989recent}
Francis Crick.
\newblock The recent excitement about neural networks.
\newblock \emph{Nature}, 337\penalty0 (6203):\penalty0 129--132, 1989.

\bibitem[Deng et~al.(2009)Deng, Dong, Socher, Li, Li, and Fei-Fei]{DengImagenet}
Jia Deng, Wei Dong, Richard Socher, Li-Jia Li, Kai Li, and Li Fei-Fei.
\newblock Imagenet: A large-scale hierarchical image database.
\newblock In \emph{2009 IEEE Conference on Computer Vision and Pattern Recognition}, pages 248--255, 2009.

\bibitem[Doersch et~al.(2015)Doersch, Gupta, and Efros]{Doersch_2015_ICCV}
Carl Doersch, Abhinav Gupta, and Alexei~A. Efros.
\newblock Unsupervised visual representation learning by context prediction.
\newblock In \emph{Proceedings of the IEEE International Conference on Computer Vision (ICCV)}, 2015.

\bibitem[Dosovitskiy et~al.(2021)Dosovitskiy, Beyer, Kolesnikov, Weissenborn, Zhai, Unterthiner, Dehghani, Minderer, Heigold, Gelly, Uszkoreit, and Houlsby]{dosovitskiy2020vit}
Alexey Dosovitskiy, Lucas Beyer, Alexander Kolesnikov, Dirk Weissenborn, Xiaohua Zhai, Thomas Unterthiner, Mostafa Dehghani, Matthias Minderer, Georg Heigold, Sylvain Gelly, Jakob Uszkoreit, and Neil Houlsby.
\newblock An image is worth 16x16 words: Transformers for image recognition at scale.
\newblock \emph{ICLR}, 2021.

\bibitem[Ghiasi et~al.(2021)Ghiasi, Cui, Srinivas, Qian, Lin, Cubuk, Le, and Zoph]{ghiasi2021simple}
Golnaz Ghiasi, Yin Cui, Aravind Srinivas, Rui Qian, Tsung-Yi Lin, Ekin~D. Cubuk, Quoc~V. Le, and Barret Zoph.
\newblock Simple copy-paste is a strong data augmentation method for instance segmentation, 2021.

\bibitem[Gomez et~al.(2017)Gomez, Ren, Urtasun, and Grosse]{gomez2017reversible}
Aidan~N Gomez, Mengye Ren, Raquel Urtasun, and Roger~B Grosse.
\newblock The reversible residual network: Backpropagation without storing activations.
\newblock \emph{Advances in neural information processing systems}, 30, 2017.

\bibitem[Grill et~al.(2020)Grill, Strub, Altch\'{e}, Tallec, Richemond, Buchatskaya, Doersch, Pires, Guo, Azar, Piot, Kavukcuoglu, Munos, and Valko]{Grillnips2020}
Jean-Bastien Grill, Florian Strub, Florent Altch\'{e}, Corentin Tallec, Pierre~H. Richemond, Elena Buchatskaya, Carl Doersch, Bernardo~Avila Pires, Zhaohan~Daniel Guo, Mohammad~Gheshlaghi Azar, Bilal Piot, Koray Kavukcuoglu, R\'{e}mi Munos, and Michal Valko.
\newblock Bootstrap your own latent a new approach to self-supervised learning.
\newblock In \emph{Proceedings of the 34th International Conference on Neural Information Processing Systems}, Red Hook, NY, USA, 2020. Curran Associates Inc.

\bibitem[He et~al.(2016)He, Zhang, Ren, and Sun]{He2016resnet}
Kaiming He, Xiangyu Zhang, Shaoqing Ren, and Jian Sun.
\newblock Deep residual learning for image recognition.
\newblock In \emph{2016 IEEE Conference on Computer Vision and Pattern Recognition (CVPR)}, pages 770--778, 2016.

\bibitem[He et~al.(2017)He, Gkioxari, Doll{\'a}r, and Girshick]{he2017mask}
Kaiming He, Georgia Gkioxari, Piotr Doll{\'a}r, and Ross Girshick.
\newblock Mask r-cnn.
\newblock In \emph{Proceedings of the IEEE international conference on computer vision}, pages 2961--2969, 2017.

\bibitem[He et~al.(2022)He, Chen, Xie, Li, Dollár, and Girshick]{He2022MAE}
Kaiming He, Xinlei Chen, Saining Xie, Yanghao Li, Piotr Dollár, and Ross Girshick.
\newblock Masked autoencoders are scalable vision learners.
\newblock In \emph{2022 IEEE/CVF Conference on Computer Vision and Pattern Recognition (CVPR)}, pages 15979--15988, 2022.

\bibitem[Jeong and Shin(2021)]{jeong2021contrad}
Jongheon Jeong and Jinwoo Shin.
\newblock Training {GAN}s with stronger augmentations via contrastive discriminator.
\newblock In \emph{International Conference on Learning Representations}, 2021.

\bibitem[Judd et~al.(2016)Judd, Albericio, Hetherington, Aamodt, and Moshovos]{judd2016stripes}
Patrick Judd, Jorge Albericio, Tayler Hetherington, Tor~M Aamodt, and Andreas Moshovos.
\newblock Stripes: Bit-serial deep neural network computing.
\newblock In \emph{Microarchitecture (MICRO), 2016 49th Annual IEEE/ACM International Symposium on}, pages 1--12. IEEE, 2016.

\bibitem[Kakogeorgiou et~al.(2022)Kakogeorgiou, Gidaris, Psomas, Avrithis, Bursuc, Karantzalos, and Komodakis]{kakogeorgiou2022attmask}
Ioannis Kakogeorgiou, Spyros Gidaris, Bill Psomas, Yannis Avrithis, Andrei Bursuc, Konstantinos Karantzalos, and Nikos Komodakis.
\newblock What to hide from your students: Attention-guided masked image modeling.
\newblock In \emph{Computer Vision -- ECCV 2022}, pages 300--318. Springer Nature Switzerland, 2022.

\bibitem[Kong and Zhang(2023)]{Kong_2023_CVPR}
Xiangwen Kong and Xiangyu Zhang.
\newblock Understanding masked image modeling via learning occlusion invariant feature.
\newblock In \emph{Proceedings of the IEEE/CVF Conference on Computer Vision and Pattern Recognition (CVPR)}, pages 6241--6251, 2023.

\bibitem[LeCun and Bengio(1998)]{lecun1998cnn}
Yann LeCun and Yoshua Bengio.
\newblock \emph{Convolutional Networks for Images, Speech, and Time Series}, page 255–258.
\newblock MIT Press, Cambridge, MA, USA, 1998.

\bibitem[Lee et~al.(2019)Lee, Lee, Han, Lee, Park, and Yoo]{lee20197}
Jinsu Lee, Juhyoung Lee, Donghyeon Han, Jinmook Lee, Gwangtae Park, and Hoi-Jun Yoo.
\newblock 7.7 lnpu: A 25.3 tflops/w sparse deep-neural-network learning processor with fine-grained mixed precision of fp8-fp16.
\newblock In \emph{2019 IEEE International Solid-State Circuits Conference-(ISSCC)}, pages 142--144. IEEE, 2019.

\bibitem[Li et~al.(2022{\natexlab{a}})Li, Zheng, Liu, Wang, Su, and Zheng]{li2022semmae}
Gang Li, Heliang Zheng, Daqing Liu, Chaoyue Wang, Bing Su, and Changwen Zheng.
\newblock Semmae: Semantic-guided masking for learning masked autoencoders.
\newblock \emph{arXiv preprint arXiv:2206.10207}, 2022{\natexlab{a}}.

\bibitem[Li et~al.(2022{\natexlab{b}})Li, Mao, Girshick, and He]{li2022exploring}
Yanghao Li, Hanzi Mao, Ross Girshick, and Kaiming He.
\newblock Exploring plain vision transformer backbones for object detection, 2022{\natexlab{b}}.

\bibitem[Lillicrap et~al.(2014)Lillicrap, Cownden, Tweed, and Akerman]{lillicrap2014random}
Timothy~P. Lillicrap, Daniel Cownden, Douglas~B. Tweed, and Colin~J. Akerman.
\newblock Random feedback weights support learning in deep neural networks, 2014.

\bibitem[Lin et~al.(2015)Lin, Maire, Belongie, Bourdev, Girshick, Hays, Perona, Ramanan, Zitnick, and Dollár]{lin2015microsoft}
Tsung-Yi Lin, Michael Maire, Serge Belongie, Lubomir Bourdev, Ross Girshick, James Hays, Pietro Perona, Deva Ramanan, C.~Lawrence Zitnick, and Piotr Dollár.
\newblock Microsoft coco: Common objects in context, 2015.

\bibitem[Lin et~al.(2017)Lin, Dollár, Girshick, He, Hariharan, and Belongie]{lin2017feature}
Tsung-Yi Lin, Piotr Dollár, Ross Girshick, Kaiming He, Bharath Hariharan, and Serge Belongie.
\newblock Feature pyramid networks for object detection, 2017.

\bibitem[Liu et~al.(2022)Liu, Gui, and Luo]{liu2022good}
Zhengqi Liu, Jie Gui, and Hao Luo.
\newblock Good helper is around you: Attention-driven masked image modeling, 2022.

\bibitem[Löwe et~al.(2020)Löwe, O'Connor, and Veeling]{löwe2020putting}
Sindy Löwe, Peter O'Connor, and Bastiaan~S. Veeling.
\newblock Putting an end to end-to-end: Gradient-isolated learning of representations, 2020.

\bibitem[Mahmoud et~al.(2020)Mahmoud, Edo, Zadeh, Awad, Pekhimenko, Albericio, and Moshovos]{mahmoud2020tensordash}
Mostafa Mahmoud, Isak Edo, Ali~Hadi Zadeh, Omar~Mohamed Awad, Gennady Pekhimenko, Jorge Albericio, and Andreas Moshovos.
\newblock Tensordash: Exploiting sparsity to accelerate deep neural network training.
\newblock In \emph{2020 53rd Annual IEEE/ACM International Symposium on Microarchitecture (MICRO)}, pages 781--795. IEEE, 2020.

\bibitem[Marblestone et~al.(2016)Marblestone, Wayne, and Kording]{marblestone2016toward}
Adam~H Marblestone, Greg Wayne, and Konrad~P Kording.
\newblock Toward an integration of deep learning and neuroscience.
\newblock \emph{Frontiers in computational neuroscience}, 10:\penalty0 94, 2016.

\bibitem[Oord et~al.(2018)Oord, Li, and Vinyals]{oord2018representation}
Aaron van~den Oord, Yazhe Li, and Oriol Vinyals.
\newblock Representation learning with contrastive predictive coding.
\newblock \emph{arXiv preprint arXiv:1807.03748}, 2018.

\bibitem[Purushwalkam and Gupta(2020)]{purushwalkam2020demystifying}
Senthil Purushwalkam and Abhinav Gupta.
\newblock Demystifying contrastive self-supervised learning: Invariances, augmentations and dataset biases, 2020.

\bibitem[Qin et~al.(2020)Qin, Samajdar, Kwon, Nadella, Srinivasan, Das, Kaul, and Krishna]{qin2020sigma}
Eric Qin, Ananda Samajdar, Hyoukjun Kwon, Vineet Nadella, Sudarshan Srinivasan, Dipankar Das, Bharat Kaul, and Tushar Krishna.
\newblock Sigma: A sparse and irregular gemm accelerator with flexible interconnects for dnn training.
\newblock In \emph{2020 IEEE International Symposium on High Performance Computer Architecture (HPCA)}, pages 58--70. IEEE, 2020.

\bibitem[Sinha et~al.(2021)Sinha, Jia, Hupkes, Pineau, Williams, and Kiela]{sinha-etal-2021-masked}
Koustuv Sinha, Robin Jia, Dieuwke Hupkes, Joelle Pineau, Adina Williams, and Douwe Kiela.
\newblock Masked language modeling and the distributional hypothesis: Order word matters pre-training for little.
\newblock In \emph{Proceedings of the 2021 Conference on Empirical Methods in Natural Language Processing}, pages 2888--2913, Online and Punta Cana, Dominican Republic, 2021. Association for Computational Linguistics.

\bibitem[Wang et~al.(2023)Wang, Song, Fan, Wang, Xie, and Zhang]{Wang_2023_CVPR}
Haochen Wang, Kaiyou Song, Junsong Fan, Yuxi Wang, Jin Xie, and Zhaoxiang Zhang.
\newblock Hard patches mining for masked image modeling.
\newblock In \emph{Proceedings of the IEEE/CVF Conference on Computer Vision and Pattern Recognition (CVPR)}, pages 10375--10385, 2023.

\bibitem[Wang and Gupta(2015)]{wang2015unsupervised}
Xiaolong Wang and Abhinav Gupta.
\newblock Unsupervised learning of visual representations using videos.
\newblock In \emph{Proceedings of the IEEE international conference on computer vision}, pages 2794--2802, 2015.

\bibitem[Wang and Qi(2022)]{wang2022contrastive}
Xiao Wang and Guo-Jun Qi.
\newblock Contrastive learning with stronger augmentations.
\newblock \emph{IEEE transactions on pattern analysis and machine intelligence}, 45\penalty0 (5):\penalty0 5549--5560, 2022.

\bibitem[Wu et~al.(2018)Wu, Xiong, Yu, and Lin]{WuCvpr2018}
Zhirong Wu, Yuanjun Xiong, Stella~X. Yu, and Dahua Lin.
\newblock Unsupervised feature learning via non-parametric instance discrimination.
\newblock In \emph{2018 IEEE/CVF Conference on Computer Vision and Pattern Recognition}, pages 3733--3742, 2018.

\bibitem[Xie et~al.(2022)Xie, Zhang, Cao, Lin, Bao, Yao, Dai, and Hu]{xie2021simmim}
Zhenda Xie, Zheng Zhang, Yue Cao, Yutong Lin, Jianmin Bao, Zhuliang Yao, Qi Dai, and Han Hu.
\newblock Simmim: A simple framework for masked image modeling.
\newblock In \emph{International Conference on Computer Vision and Pattern Recognition (CVPR)}, 2022.

\bibitem[Xiong et~al.(2020)Xiong, Ren, and Urtasun]{xiong2020loco}
Yuwen Xiong, Mengye Ren, and Raquel Urtasun.
\newblock Loco: Local contrastive representation learning.
\newblock In \emph{Proceedings of the 34th International Conference on Neural Information Processing Systems}, Red Hook, NY, USA, 2020. Curran Associates Inc.

\bibitem[Yang et~al.(2020)Yang, Ghasemazar, Ren, Golub, Lemieux, and Lis]{yang2020procrustes}
Dingqing Yang, Amin Ghasemazar, Xiaowei Ren, Maximilian Golub, Guy Lemieux, and Mieszko Lis.
\newblock Procrustes: a dataflow and accelerator for sparse deep neural network training.
\newblock In \emph{2020 53rd Annual IEEE/ACM International Symposium on Microarchitecture (MICRO)}, pages 711--724. IEEE, 2020.

\bibitem[Zhang et~al.(2019)Zhang, Chen, Song, and Li]{zhang2019eager}
Jiaqi Zhang, Xiangru Chen, Mingcong Song, and Tao Li.
\newblock Eager pruning: algorithm and architecture support for fast training of deep neural networks.
\newblock In \emph{2019 ACM/IEEE 46th Annual International Symposium on Computer Architecture (ISCA)}, pages 292--303. IEEE, 2019.

\bibitem[Zhang et~al.(2022)Zhang, McDanel, and Kung]{zhang2022fast}
Sai~Qian Zhang, Bradley McDanel, and HT Kung.
\newblock Fast: Dnn training under variable precision block floating point with stochastic rounding.
\newblock In \emph{2022 IEEE International Symposium on High-Performance Computer Architecture (HPCA)}, pages 846--860. IEEE, 2022.

\bibitem[Zhang et~al.(2023)Zhang, Tambe, Cuevas, Wei, and Brooks]{zhang2023camel}
Sai~Qian Zhang, Thierry Tambe, Nestor Cuevas, Gu-Yeon Wei, and David Brooks.
\newblock Camel: Co-designing ai models and embedded drams for efficient on-device learning.
\newblock \emph{arXiv preprint arXiv:2305.03148}, 2023.

\end{thebibliography}
}


\end{document}